\documentclass[12pt]{article}

\usepackage{amsmath,amssymb}
\usepackage{graphicx}
\usepackage{caption}
\usepackage{color}
\usepackage{dcolumn}
\usepackage{bm}
\usepackage[numbers,super,comma,sort&compress]{natbib}
\usepackage{float}
\usepackage{hyperref}
\hypersetup{colorlinks=true,citecolor=black,urlcolor=blue}
\definecolor{background-color}{gray}{0.98}

\renewcommand{\footnotemark}{}
\renewcommand{\baselinestretch}{1}

\DeclareMathOperator*{\median}{median}

\DeclareMathOperator*{\minimize}{minimize}
\def\bmu{\boldsymbol{\mu}}
\def\hbmu{\hat{\boldsymbol{\mu}}}

\def\bu{{\boldsymbol{u}}}
\def\bx{{\boldsymbol{x}}}

\def\bX{{\boldsymbol{X}}}
\def\bT{{\boldsymbol{T}}}
\def\bS{{\boldsymbol{S}}}

\def\bSigma{\boldsymbol{\Sigma}}

\def\MAD{\text{MAD}}

\def\LTS{\text{LTS}}

\title{Anomaly Detection by Robust Statistics}

\author{Peter J. Rousseeuw and Mia Hubert\thanks{Department
 of Mathematics, KU Leuven, Celestijnenlaan 200B, BE-3001 
Leuven, Belgium.}}

\date{October 14, 2017}
\begin{document}
\maketitle

\thispagestyle{empty}

\begin{center}

\subsubsection*{Abstract}
\begin{flushleft}
Real data often contain anomalous cases, also known as 
outliers. These may spoil the resulting analysis but 
they may also contain valuable information.
In either case, the ability to detect such anomalies
is essential.
A useful tool for this purpose is robust statistics, which
aims to detect the outliers by first fitting the majority 
of the data and then flagging data points that deviate 
from it.
We present an overview of several robust methods and 
the resulting graphical outlier detection tools. 
We discuss robust procedures for univariate, 
low-dimensional, and high-dimensional data, 
such as estimating location and scatter, linear 
regression, principal component analysis, classification, 
clustering, and functional data analysis.
Also the challenging new topic of cellwise outliers is
introduced.
\end{flushleft}
\end{center}

\clearpage
 
\makeatletter
\renewcommand\@biblabel[1]{#1.}
\makeatother
\bibliographystyle{unsrtnat} % This is WIREs `Vancouver style'
\renewcommand{\baselinestretch}{1.5}
\normalsize

\clearpage

\section*{\sffamily \Large INTRODUCTION} 
In real data sets it often happens that some 
cases behave differently from the majority. 
Such data points are called {\it anomalies} in 
machine learning, and {\it outliers} in statistics. 
Outliers may be caused by errors, but they could
also have been recorded under exceptional 
circumstances, or belong to another population.  
It is very important to be able to detect anomalous
cases, which may (a) have a harmful effect on the 
conclusions drawn from the data, or (b) contain 
valuable nuggets of information.

In practice one often tries to detect outliers using 
diagnostics starting from a classical fitting method. 
However, classical methods can be affected by outliers 
so strongly that the resulting fitted model may not 
allow to detect the deviating observations.
This is called the {\it masking} effect. Additionally, 
some good data points might even appear to be outliers, 
which is known as {\it swamping}. To avoid these 
effects, the goal of robust statistics is to 
find a fit which is close to the fit we would have 
found without the outliers. We can then identify the 
outliers by their large `deviation' (e.g. its distance 
or residual) from that robust fit.

First we describe some robust procedures for detecting
anomalies in univariate location and scale, as well as 
in multivariate data and in the linear regression
setting.
For more details on this part 
see~\cite{Hubert:ReviewHighBreakdown,
Rousseeuw:RobReg,Maronna:RobStat}.
Next we discuss principal component analysis (PCA) and
some available robust methods for classification,
clustering, and functional data analysis. Finally we 
introduce the emerging 
research topic of detecting cellwise anomalies.

\section*{\sffamily \Large ESTIMATING UNIVARIATE
                           LOCATION AND SCALE}
As an example of univariate data, suppose we have five
measurements of a length:
\begin{equation}\label{eq:cleanunivar}
   6.27,\;\;\; 6.34,\;\;\; 6.25,\;\;\; 6.31,\;\;\; 6.28
\end{equation}
and we want to estimate its true value. For this,
one usually computes the sample mean
$\bar{x}=\frac{1}{n} \sum_{i=1}^n x_i\;$
which in this case equals 
$\bar{x} = (6.27 + 6.34 + 6.25 + 6.31 + 6.28)/5 = 6.29\;$. 
Let us now suppose that the fourth measurement has been
recorded wrongly and the data become
\begin{equation}\label{eq:contunivar}
   6.27,\;\;\; 6.34,\;\;\; 6.25,\;\;\; 63.1,\;\;\; 6.28\;.
\end{equation}
In this case we obtain $\bar{x}= 17.65$, which is far 
off. Alternatively, we could
also compute the {\it median} of these data. For this we
sort the observations in \eqref{eq:contunivar} from 
smallest to largest:
\begin{equation*}
   6.25 \leqslant 6.27 \leqslant 6.28 \leqslant
	 6.34 \leqslant 63.10\;\;.
\end{equation*}
The median is the middle value, here yielding 6.28,
which is still reasonable. We say that the median is
more robust against an outlier.

More generally, the location-scale model states that the 
$n$ univariate observations $x_i$ are independent and 
identically distributed (i.i.d.) with distribution 
function 
$F((x-\mu)/\sigma)$ where $F$ is known. Typically $F$ is 
the standard gaussian distribution function $\Phi$. 
We then want to find estimates for the unknown center
$\mu$ and the unknown scale parameter $\sigma$.

The classical estimate of location is the mean.
As we saw above, the mean is very sensitive to aberrant 
values among the $n$ observations. 
In general, replacing even a single observation by a 
very large value can change the mean completely. 
We say that the 
\emph{breakdown value}~\cite{Hampel:Robust,Donoho:BDP} 
of the sample mean is $1/n$, so it becomes $0\%$ for 
large $n$. 
In general, the breakdown value is the smallest 
proportion of observations in the data set that need to 
be replaced to carry the estimate arbitrarily far away. 
A breakdown value of 0\% is thus the worst possible. 
See~\cite{Hubert:WIRE-break} for precise definitions 
and extensions.												
The robustness of an estimator is also measured by 
its \emph{influence function}~\cite{Hampel:IFapproach} 
which measures the effect of a single outlier. 
The influence function of the mean is unbounded, which 
again illustrates that the mean is not robust.

For the general definition of the median, we denote 
the $i$th ordered observation as $x_{(i)}$. 
The median is defined as $x_{((n+1)/2)}$ if $n$ is odd 
and $(x_{(n/2)}+x_{(n/2+1)})/2$ if $n$ is even. 
Its breakdown value is about $50\%$, meaning that the 
median can resist almost $50\%$ of outliers. This is
the best possible breakdown value since the clean data 
need to be in the majority.

The situation for the scale parameter $\sigma$ is similar.
The classical estimator is the \emph{standard deviation} 
$s=\sqrt{\sum_{i=1}^n (x_i - \bar{x})^2/(n-1)}.$ 
Since a single outlier can already make $s$ arbitrarily 
large, its breakdown value is $0\%$. 
For instance, for the clean data \eqref{eq:cleanunivar} 
above we have $s = 0.035$, whereas for the data 
\eqref{eq:contunivar}
with the outlier we obtain $s = 25.41$\;!

A robust measure of scale is the {\bf m}edian of all
{\bf a}bsolute {\bf d}eviations from the median (MAD), 
given by 
\begin{equation}
  \MAD = 1.4826 \, \median_{i=1,\dots,n}
	          |x_i - \median_{j=1,\dots,n}(x_j)|\;\;.
  \label{eq:MAD}
\end{equation}
The constant 1.4826 is a correction factor which makes 
the MAD consistent at gaussian distributions. 
The MAD of \eqref{eq:contunivar} is the same as that 
of \eqref{eq:cleanunivar}, namely 0.044.
We can also use the
$Q_n$ estimator~\cite{Rousseeuw:scale}, defined as
\begin{equation*}
   Q_n = 2.2219 \, \{|x_i-x_j|; i < j \}_{(k)}
\end{equation*}
with $k=\binom{h}{2} \approx \binom{n}{2}/4$ 
and $h=\lfloor \frac{n}{2} \rfloor +1$. 
Here $\lfloor \ldots \rfloor$ rounds down to the nearest 
integer. 
This scale estimator is thus the first quartile of all 
pairwise distances between two data points. The breakdown 
value of both the MAD and the $Q_n$ estimator is $50\%$.

Also the (normalized) interquartile range (IQR) can be 
used, given by $\text{IQR}= 0.7413(Q_3 - Q_1)$
where $Q_1 = x_{\lfloor n/4 \rfloor}$ is the first 
quartile of the data and $Q_3 = x_{\lceil 3n/4 \rceil}$ 
is the third quartile.
The IQR has a simple expression but its breakdown value 
is only $25\%$, so it is less robust than the MAD 
and $Q_n$.

The robustness of the median (and the MAD) comes at a 
price: 
at the gaussian model it is less efficient than the mean. 
Many robust procedures have been proposed that strike
a balance between robustness and efficiency, such as 
location M-estimators~\cite{Huber:Robloc}. They are 
defined implicitly as the solution of the equation
\begin{equation}\label{eq:huberm}
\sum_{i=1}^n \psi\Bigl(\frac{x_i - \hat{\mu}}
               {\hat{\sigma}}\Bigr) = 0
\end{equation}
for a given real function $\psi$\;. The
denominator $\hat{\sigma}$ is an initial robust scale 
estimate such as $Q_n$\;. A solution $\hat{\mu}$ 
to~\eqref{eq:huberm} can be found by
an iterative algorithm, starting from the initial location
estimate $\hat{\mu}^{(0)} = \median_i (x_i)$. 
Popular choices for $\psi$ are the Huber function 
$\psi(x)= x \min(1,c/|x|)$ and Tukey's bisquare function
$\psi(x)= x (1-(x/c)^2)^2 I(|x| \leqslant c)$. 
These M-estimators contain a tuning parameter $c$ which 
needs to be chosen in advance. Also M-estimators for
the scale parameter $\sigma$ exist.

People often use rules to detect outliers. The classical 
rule is based on the $z$-scores of the observations, 
given by
\begin{equation}\label{eq:zscore}
    z_i = \frac{x_i-\bar{x}}{s}
\end{equation}
where $s$ is the standard deviation of the data.
More precisely, the rule flags $x_i$ as outlying 
if $|z_i|$ exceeds 2.5, say.
But in the above example \eqref{eq:contunivar} with the
outlier, the $z$-scores are 
   $$ -0.45,\;\;\; -0.45,\;\;\; -0.45,\;\;\;
	   1.79,\;\;\; -0.45 $$
so none of them attains 2.5. The largest value is only
1.79, which is quite similar to the largest $z$-score for
the clean data \eqref{eq:cleanunivar}, which equals 1.41. 
The $z$-score of the outlier is small because it subtracts 
the nonrobust mean (which was drawn toward the outlier) 
and because it divides by the nonrobust standard deviation
(which the outlier has made much larger than in the
clean data).
Plugging in robust estimators of location and scale such 
as the median and the MAD yields the robust scores
\begin{equation}\label{eq:robzscore}
   \frac{x_i-\median_j(x_j)}{\text{MAD}_j(x_j)}
\end{equation}	
which yield a much more reliable outlier detection tool.
Indeed, in the contaminated example \eqref{eq:contunivar} 
the robust scores are
$$ -0.22,\;\;\; 1.35,\;\;\; -0.67,\;\;\;
   1277.5,\;\;\; 0.0 $$
where that of the outlier greatly exceeds the 2.5 
cutoff.
 
Also Tukey's boxplot is often used to pinpoint possible 
outliers. In this plot a box is drawn from the first 
quartile $Q_1$ of the data to the third quartile $Q_3$\;. 
Points outside the interval 
$[Q_1-1.5\ \text{IQR},Q_3+1.5\ \text{IQR}]$,
called the {\it fence}, are traditionally marked as outliers. 
Note that the boxplot assumes symmetry, since we add the
same amount to $Q_3$ as what we subtract from $Q_1$. 
At asymmetric distributions the usual boxplot typically 
flags many regular data points as outliers. 
The skewness-adjusted boxplot~\cite{Hubert:AdjBoxplot}
corrects for this by using a robust measure of 
skewness~\cite{Brys:Medcouple}
in determining the fence.

\section*{\sffamily \Large MULTIVARIATE LOCATION AND
                           COVARIANCE ESTIMATION} 
From now on we assume that the data are $d$-dimensional 
and are stored in an $n \times d$ data
matrix $\bX=(\bx_1,\ldots,\bx_n)^T$ 
with $\bx_i=(x_{i1},\ldots,x_{id})^T$ the $i$th data point. 
Classical measures of location and scatter are given by 
the empirical mean
   $\bar{\bx}=\frac{1}{n} \sum_{i=1}^n \bx_i$
and the empirical covariance matrix
   $\bS_\bX=\sum_{i=1}^n 
	  (\bx_i - \bar{\bx})(\bx_i-\bar{\bx})^T/(n-1)$.
As in the univariate case, both classical estimators have 
a breakdown value of $0\%$, that is, a small fraction of 
outliers can completely ruin them.

\begin{figure}[H]
\begin{center}
\includegraphics[keepaspectratio=true,scale=1.0]
                {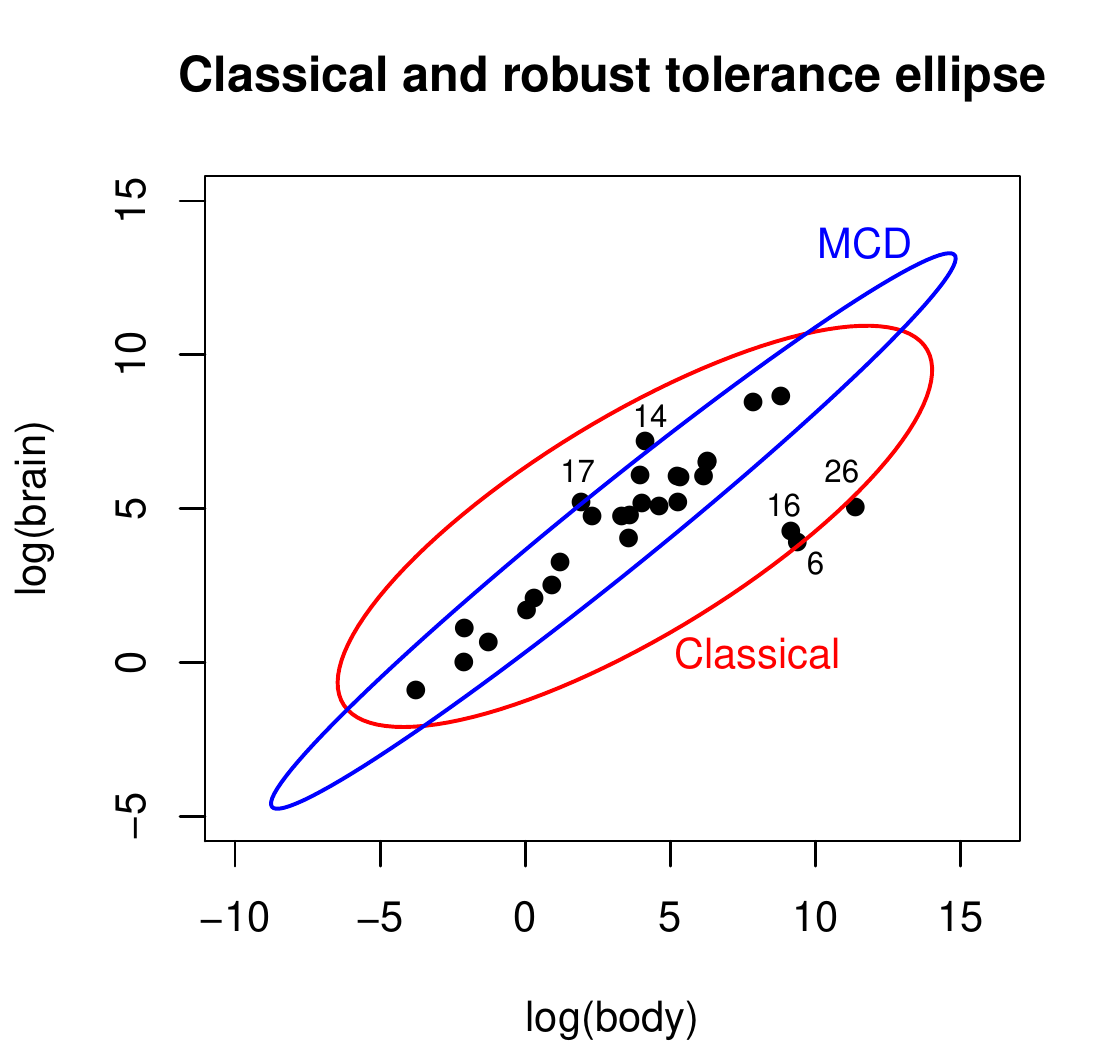}	
\caption{Animal data: tolerance ellipse of the classical
         mean and covariance matrix (red), and that of 
				 the robust location and scatter matrix (blue).}
\label{brainlog_ellipses} 
\end{center}
\end{figure}

As an illustration we consider a bivariate dataset 
(from page 59 in \cite{Rousseeuw:RobReg}) containing
the logarithms of body weight and brain weight of 28 animal
species, with scatterplot in Figure~\ref{brainlog_ellipses}.
Any point $\bx$ has a so-called Mahalanobis distance (or
`generalized distance')
\begin{equation}
  \text{MD}(\bx,\hbmu,\hat{\bSigma})=
  \sqrt{(\bx-\hbmu)^T\hat{\bSigma}^{-1}(\bx-\hbmu)}
\label{eq:RD}
\end{equation}
to the mean $\hbmu=\bar{\bx}$\;, taking the covariance
matrix $\hat{\bSigma}=\bS_\bX$ into account. 
The $\text{MD}$ is constant on ellipsoids. The so-called
97.5\% tolerance ellipsoid is given by
$MD(\bx) \leqslant \sqrt{\chi_{d,0.975}^2}$ where
$\chi_{d,0.975}^2$ is the 0.975 quantile of the 
chi-squared distribution with $d$ degrees of freedom.
In this bivariate example $d=2$, and the resulting ellipse
is drawn in red. 
We see that it is inflated in the direction of the three
outliers 6, 16, and 26 which are dinosaurs having low brain
weight and high body weight.
As a result these data points fall near the boundary of 
the tolerance ellipse, i.e. their $MD(\bx_i)$ are not very 
high.

Alternatively we can compute robust estimates of location 
and scatter (covariance), for instance by 
the {\it Minimum Covariance Determinant} (MCD) 
method~\cite{Rousseeuw:LMS,Rousseeuw:MVE}. The MCD
looks for those $h$ observations in the data set (where the 
number $h$ is given by the user) whose classical covariance 
matrix has the lowest possible determinant. 
The MCD estimate of location $\hbmu$ is then the average 
of these $h$ points, whereas the MCD estimate of scatter 
$\hat{\bSigma}$ is their covariance matrix, multiplied by
a consistency factor.
(By default this is then followed by a reweighting step to
improve efficiency at gaussian data.)
Instead of Mahalanobis distances we can then compute
robust distances, again given by~\eqref{eq:RD} but now
with the robust estimates $\hbmu$ and $\hat{\bSigma}$\;.
This yields the robust tolerance ellipse shown in blue
in Figure~\ref{brainlog_ellipses}. 
This ellipse exposes the three dinosaurs, and we see
two species near the upper boundary, 17 (rhesus monkey)
and 14 (human).

\begin{figure}[H]
\begin{center}
\includegraphics[keepaspectratio=true,scale=1.0]
                {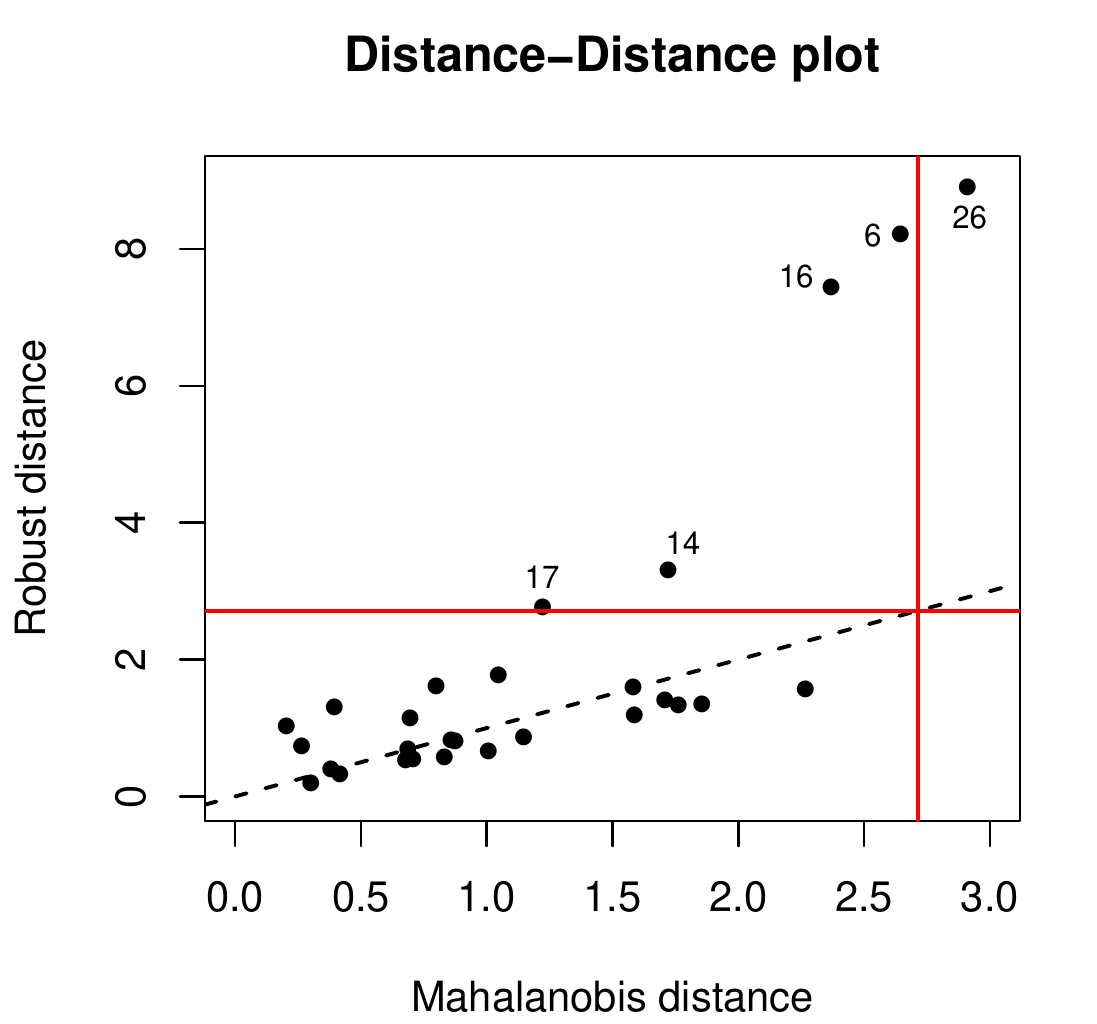}	
\caption{Animal data: Robust distance versus classical
         Mahalanobis distance.}
\label{brainlog_DD} 
\end{center}
\end{figure}

In dimension $d=4$ or higher it becomes infeasible to
visualize the tolerance ellipsoid, but we still have the
distances. The {\it distance-distance plot} (DD-plot) in 
Figure~\ref{brainlog_DD} shows the robust distance 
$\text{RD}(\bx_i)$ of each data point versus its
classical Mahalanobis distance $\text{MD}(\bx_i)$\;.
The horizontal and vertical cutoff lines are at
$\sqrt{\chi_{d,0.975}^2}$ and the dashed line is where
classical and robust distances coincide.
We see that the $RD(\bx_i)$ flag all the outliers in this
dataset, while the $MD(\bx_i)$ don't. 
For a dataset in which they are
very similar we can trust classical statistical methods,
but when they differ much (like here) the DD-plot detects
the outlying data points. This does not imply we should
somehow delete them, but rather that they should be
investigated and understood. Outliers are not necessarily
`errors': they can also correspond to unusual 
circumstances or be members of a different population.

The MCD estimator, as well as its weighted version, has a 
bounded influence function and breakdown value $(n-h+1)/n$, 
hence the number $h$ determines the robustness of the 
estimator. The MCD has its highest possible breakdown 
value when $h = \lfloor (n+p+1)/2 \rfloor$. 
When a large proportion of contamination is 
expected, $h$ should thus be chosen close to $0.5n$.
Otherwise an intermediate value for $h$, such as $0.75n$, 
is recommended to obtain a higher finite-sample efficiency. 
Reference~\cite{Hubert:WIRE-MCD} gives a more detailed
overview of the MCD estimator and its properties.

The computation of the MCD estimator is non-trivial and 
naively requires an exhaustive investigation of 
all $h$-subsets out of $n$. 
Fortunately a much faster algorithm was constructed,
called FastMCD~\cite{Rousseeuw:FastMCD}. 
It starts by randomly drawing many $p+1$ observations from 
the data set. Based on these subsets, $h$-subsets are 
obtained by means of so-called $C$-steps 
(see~\cite{Rousseeuw:FastMCD} for details). 
More recently an even faster algorithm called DetMCD was 
devised~\cite{DetMCD} which carries out a deterministic
computation instead of random sampling.

The MCD assumes that $n > d$, so there must be more data
points than dimensions, and it works best when $n > 5d$.
When there are more than, say, 20 dimensions and/or
$d \geqslant n$ other methods are needed. 
One is to compute
robust principal components as described in a section below.
Another is to use the {\it minimum regularized covariance
determinant} (MRCD) method~\cite{MRCD}. This approach
minimizes $\text{det}\{\rho \bT + (1-\rho)\bS_H\}$ where
$\bT$ is a positive definite target matrix and $\bS_H$ is
the covariance matrix of a subset $H$ with $h$ data points.
The combined matrix is always positive definite, whereas 
$\text{det}\{\bS_H\} = 0$ when $d \geqslant n$.

Many other robust estimators of location and scatter have 
been presented in the literature. The first such estimator 
was proposed by Stahel~\cite{Stahel:SDest} and 
Donoho~\cite{Donoho:Depth} (see also~\cite{Maronna:SDEst}).
They defined the so-called Stahel-Donoho outlyingness of 
a data point $\bx_i$ as
\begin{equation}
  \text{outl}({\bx_i}) = \max_{\bu}
   \frac{|\bx_i^T\bu - \median_{j=1,\dots,n}
	   (\bx_j^T\bu)|}
     {\MAD_{j=1,\dots,n}(\bx_j^T\bu)}
  \label{eq:staheldonoho}
\end{equation}
where the maximum is over all directions (i.e., all 
$d$-dimensional unit length vectors $\bu$), 
and $\bx_j^T\bu$ 
is the projection of $\bx_j$ on the direction $\bu$.
In each direction this uses the robust $z$-scores
\eqref{eq:robzscore}.
Recently a version of \eqref{eq:staheldonoho} suitable 
for skewed distributions was proposed~\cite{DOIV}.

Multivariate M-estimators~\cite{Maronna:Mest} have a 
low breakdown value due to possible implosion 
of the estimated scatter matrix.  
More recent robust estimators of multivariate location 
and scatter with high breakdown value include 
S-estimators~\cite{Rousseeuw:RobReg, 
Davies:AsymptSest}, MM-estimators~\cite{Tatsuoka:MMest}, 
and the OGK estimator~\cite{Maronna:OGK}.

\section*{\sffamily \Large LINEAR REGRESSION} 
The multiple linear regression model assumes that there
are $d$ `explanatory' $x$-variables as well as a response
variable $y$ which can be approximated by a linear 
combination of the $x$-variables. More precisely, the 
model says that for all data points $(\bx_i,y_i)$ it 
holds that
\begin{equation}
y_i=\beta_{0}+\beta_1 x_{i1}+\dots+\beta_d x_{id}+
  \varepsilon_i 
  \qquad i=1,\dots,n \label{eq:multreg}
\end{equation}
where the errors $\varepsilon_i$ are assumed to be 
independent and identically distributed with zero 
mean and constant variance $\sigma^2$. 
Applying a regression estimator to the data yields 
$d+1$ regression coefficients, combined as
$\boldsymbol{\hat{\beta}}=
 (\hat{\beta}_0,\dots,\hat{\beta_{d}})^T$.
The residual $r_i$ of case $i$ is defined as the 
difference between the observed response $y_i$ and 
its estimated value $\hat{y}_i$\;.
 
The classical least squares (LS) method to estimate 
$\boldsymbol{\beta}$
minimizes the sum of the squared residuals. 
It is popular because it allows to compute the 
regression estimates explicitly, and it is optimal 
if the errors have a gaussian distribution.
Unfortunately LS is extremely sensitive to
outliers, i.e.\ data points that do not obey the 
linear pattern formed by the majority of the data.

\begin{figure}[H]
\begin{center}
\includegraphics[keepaspectratio=true,scale=1.0]
                {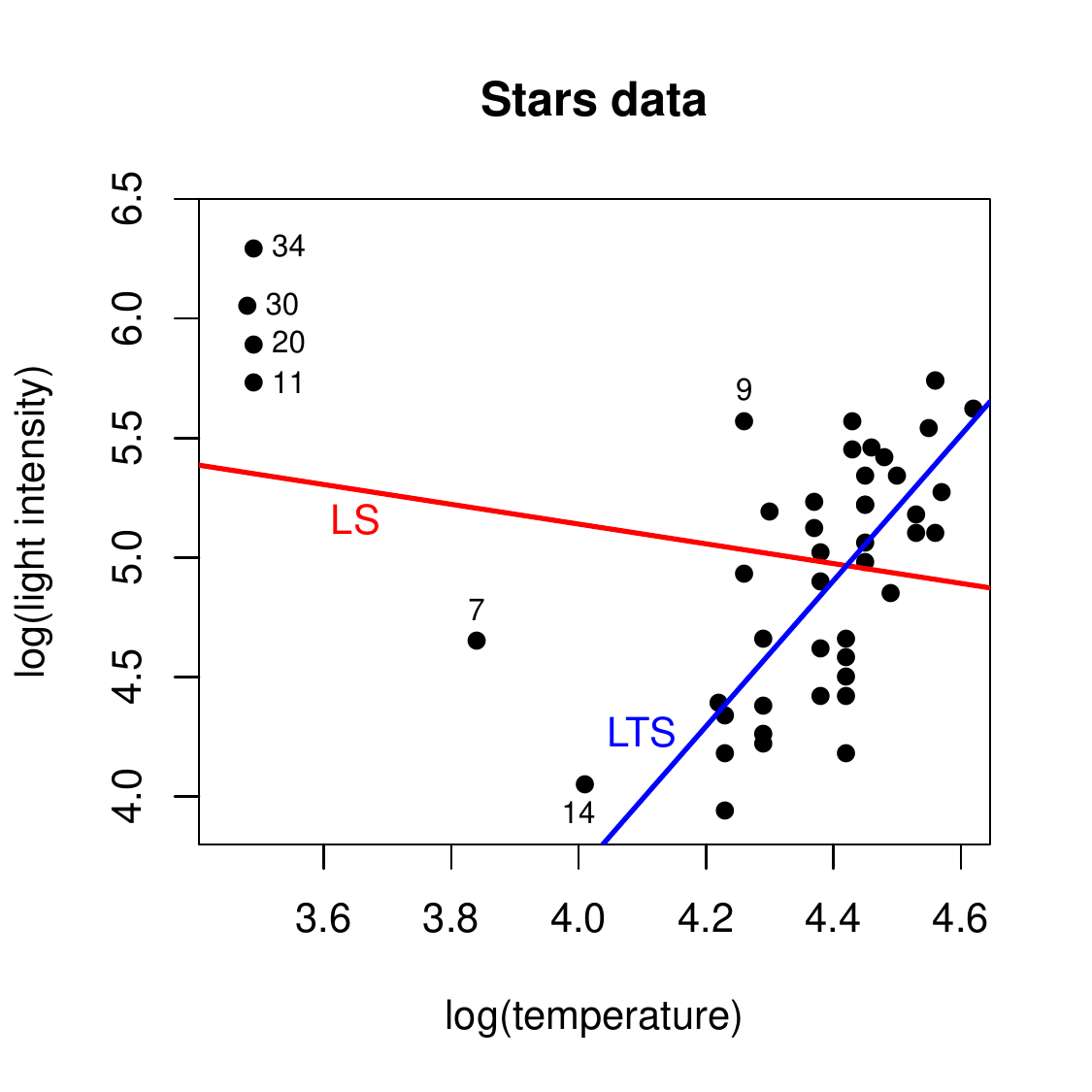}	
\caption{Stars data: Classical least squares line 
         (red) and robust line (blue).}
\label{stars_lines} 
\end{center}
\end{figure}

For instance, Figure~\ref{stars_lines} shows the 
Hertzsprung-Russell diagram of the star 
cluster CYG OB1,
containing 47 stars. The $x$-coordinate of each star is
the logarithm of its surface temperature, and the
$y$-coordinate is the logarithm of its light intensity.
Most of the stars belong to the so-called main
sequence, whereas 11, 20, 30, 34 are giant stars
and 7 is intermediate. The least squares line is shown
in red, and has a negative slope although the main
sequence slopes upward. It has been pulled away by
the leverage exerted by the four giant stars. 
As an unfortunate side effect, the giant stars do not 
have larger absolute residuals than some of the main
sequence stars, so only looking at residuals would
not allow to detect them.

The blue line on the other hand is the result of
a robust regression method, the {\it Least Trimmed 
Squares} (LTS) estimator proposed 
by Rousseeuw~\cite{Rousseeuw:LMS}. 
The LTS is given by
\begin{equation}
  \minimize_{\boldsymbol{\beta}} \;\;
  \sum_{i=1}^h (r^2)_{(i)} \label{eq:LTS}
\end{equation}
where $(r^2)_{(1)} \leqslant (r^2)_{(2)}
 \leqslant \dots \leqslant (r^2)_{(n)}$ are the 
ordered squared residuals. 
(They are first squared, and then ordered.) 
By not adding {\it all} the squared residuals the LTS 
makes it possible to fit the majority of the data, 
whereas the outliers can have large residuals. 
In Figure \ref{stars_lines} the blue line indeed fits
the main sequence stars, and stays far from the four
giant stars so the latter will have large residuals
from that line. (Note that the giant stars are not 
`errors' but correct observations of
members of a different population.)

The value $h$ in \eqref{eq:LTS} plays the same role as 
in the MCD estimator. For $h \approx n/2$ we find a
breakdown value of $50\%$, whereas for larger $h$ we 
obtain roughly $(n-h)/n$. A fast algorithm for the LTS 
estimator (FAST-LTS) has been 
developed~\cite{Rousseeuw:FastLTS}.
The scale of the errors $\sigma$ can be estimated by
$\hat{\sigma}_{\LTS}^2
 =c_{h,n}^2\sum^{h}_{i=1}(r^2)_{(i)}/h$
where $r_i$ are the residuals from the LTS f\/it, and 
$c_{h,n}$ is a constant that makes $\hat{\sigma}_{\LTS}$ 
consistent at gaussian error distributions,
as described in~\citep{Pison:chn}.
We can then identify outliers by their large 
standardized LTS residuals $r_i/\hat{\sigma}_{\LTS}$.
We can also use the standardized LTS residuals to assign
a weight to every observation. The weighted LS estimator 
with these LTS weights inherits the nice robustness 
properties of LTS, but is more efficient and yields all 
the usual inferential output such as t-statistics, 
F-statistics, an $R^2$ statistic, and the corresponding 
$p$-values. Alternatively, inference for LTS can be 
based on the fast robust bootstrap proposed 
in~\citep{Willems:FRB,Matias:FRB}.

\begin{figure}[H]
\begin{center}
\includegraphics[keepaspectratio=true,scale=1.0]
                {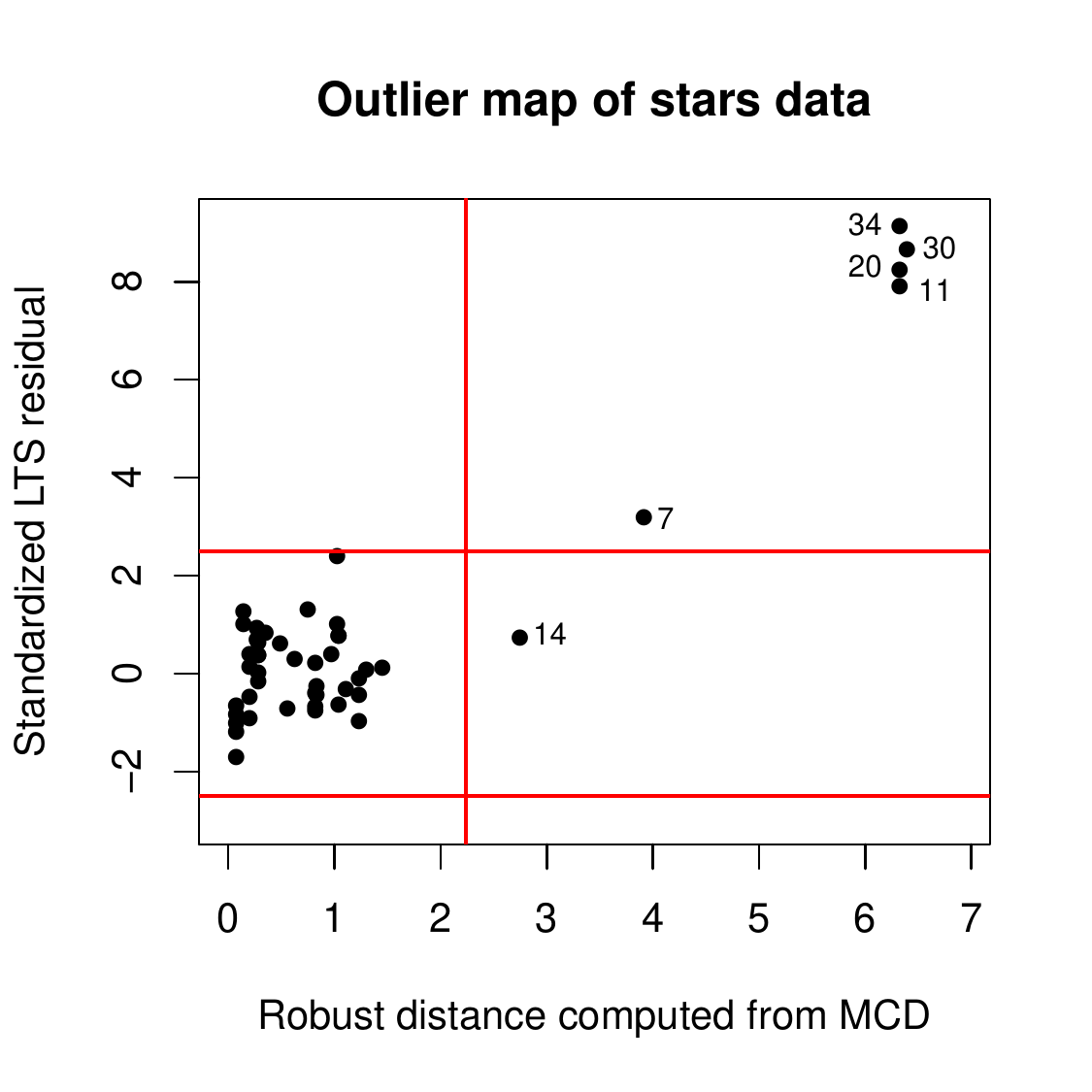}	
\caption{Stars data: Standardized robust residuals
         of $y$ versus robust distances of $x$.}
\label{stars_LTS_OM} 
\end{center}
\end{figure}

In most situations we have more than one explanatory
variable, and for dimension $d = 3$ and higher it is 
no longer possible to perceive the linear patterns
by eye. It is in those cases that robust regression
becomes the most useful.  
To flag and interpret the outliers we can use the 
{\it outlier map} of~\cite{Rousseeuw:Diagnostic} 
which plots the standardized LTS residuals versus 
robust distances~\eqref{eq:RD} based on (for instance)
the MCD estimator applied to the $x$-variables only.
Figure~\ref{stars_LTS_OM} is the outlier map of the
stars data. The tolerance band on the standardized
LTS residuals is given by the horizontal lines
at $2.5$ and $-2.5$\,, and the vertical line
corresponds to the cutoff
value $\sqrt{\chi_{d,0.975}^2}$ on the robust
distances of the $\bx_i$.
Data points $(\bx_i,y_i)$ whose residuals fall outside
the horizontal tolerance band are called
{\it regression outliers}.
On the other hand, data points $(\bx_i,y_i)$ whose
robust distance $RD(\bx_i)$ exceeds the cutoff are
called {\it leverage points}, irrespective of their
response $y_i$.
So, the outlier map diagnoses 4 types of data points.
Those with small $|r_i|$ and small $RD(\bx_i)$ are
considered {\it regular observations}, and most points
in Figure~\ref{stars_LTS_OM} fall in that rectangle.
Those with large residuals $r_i$ (positive or negative)
and small $RD(\bx_i)$ are called {\it vertical
outliers} (there are none in this figure).
Those with small $|r_i|$ but large $RD(\bx_i)$ (like
point 14) are called {\it good leverage points} because 
they improve the accuracy of the fit.
And finally, regression outliers that are also
leverage points are called {\it bad leverage points},
like the 4 giant stars in this example. Note that the
outlier map permits nuanced statements, for instance
point 7 is a leverage point but only slightly bad.

\begin{figure}[H]
\begin{center}
\includegraphics[width=1.0\textwidth]
                {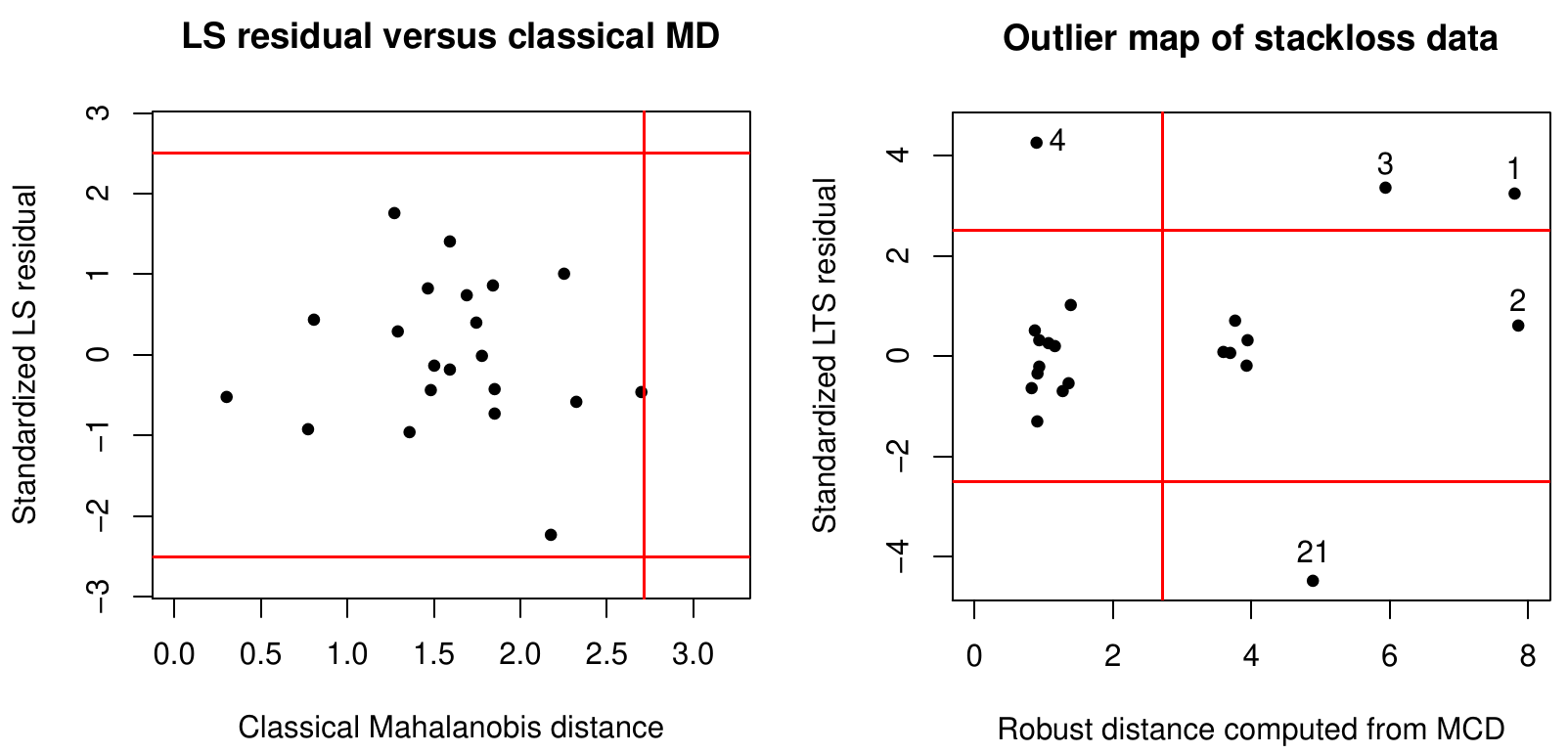}						
\caption{Stackloss data: (left) standardized nonrobust 
         LS residuals of $y$ versus nonrobust distances 
				 of $x$; (right) same with robust residuals and 
				 robust distances.}	
\label{stackloss_LS_LTS_OM} 
\end{center}
\end{figure}

The main benefit of the outlier map is when the data
has more dimensions. For instance, the stackloss 
data~\cite{Brownlee}
is a benchmark data set with 21 points with $d=3$
explanatory variables, an intercept term and a
response variable $y_i$. We cannot easily interpret
such 4-dimensional data, but we can still look at the
outlier map in the right panel of 
Figure~\ref{stackloss_LS_LTS_OM}. 
We see that 4 is a vertical
outlier, 1, 3, and 21 are bad leverage points, and 2
is a good leverage point. 
Note that the left panel of 
Figure~\ref{stackloss_LS_LTS_OM} does not flag any of
these points because it uses the classical LS
residuals and the classical distances $MD(\bx_i)$,
both of which tend to mask atypical points.

It should be stressed that flagging atypical points 
with the outlier map (as in the right panel of
Figure~\ref{stackloss_LS_LTS_OM}) is not the end of 
the analysis, but rather a new start.
The next step should be to try to figure out why 
these points are atypical and/or to improve the model
by things like data transformation, model 
selection, higher order terms, etc.
For variance selection in robust regression
see~\cite{Khan:fwd}.
When the dimension is very high one needs to resort
to sparse methods, for instance by penalization. 
The first sparse methods for robust regression were 
developed in~\citep{RLARS,SparseLTS}.

Historically, the earliest attempts at robust 
regression were least 
absolute deviations (LAD, also called $L^1$), 
M-estimators~\cite{Huber:RobStat},
R-estimators~\cite{Jureckova:Rest}, and
L-estimators~\cite{Koenker:Lest}.
The breakdown value of all these methods is $0\%$ 
because of their vulnerability to bad leverage points. 
Generalized M-estimators (GM-estimators)
\cite{Hampel:IFapproach} were the first to attain a 
positive breakdown value, which unfortunately still 
went down to zero for increasing $p$.

The low finite-sample efficiency of LTS can be improved 
by replacing its objective function by a more efficient 
scale estimator applied to the residuals $r_i$. 
This approach has led to the introduction of 
high-breakdown regression 
S-estimators~\cite{Rousseeuw:Sest} and 
MM-estimators~\cite{Yohai:MMest}.

\section*{\sffamily \Large PRINCIPAL COMPONENT ANALYSIS}
Principal component analysis (PCA) is a popular 
dimension reduction method. It tries to explain the 
covariance structure of the data by means of a 
(hopefully small) number of components. 
These components are linear combinations of the original 
variables, and often allow for an interpretation and a 
better understanding of the different sources of 
variation. 
PCA is often the first step of the data analysis, 
followed by other multivariate techniques.

In the classical approach, the first principal component 
corresponds to the direction in which the projected 
data points have the largest variance. The second component 
is then taken orthogonal to the first and must again
maximize the variance of the data points projected on it
(subject to the orthogonality constraint). 
Continuing in this way produces all the principal 
components.
It turns out that the classical principal components 
correspond to the eigenvectors of the empirical covariance 
matrix. Unfortunately, both the classical variance (which 
is being maximized) and the classical covariance matrix 
(which is being decomposed) are very sensitive to anomalous 
observations. 
Consequently, the first components from classical PCA are 
often attracted towards outlying points, and may not 
capture the variation of the regular observations.

A first group of robust PCA methods is obtained by 
replacing the classical covariance matrix by a robust 
covariance estimator, such as the weighted MCD estimator 
or MM-estimators~\cite{Croux:PCAMCD,Salibian:MMPCA}.
Unfortunately the use of these covariance estimators is 
limited to small to moderate dimensions since they are 
not defined when $d \geqslant n$.

A second approach to robust PCA uses 
{\it Projection Pursuit} techniques. These methods 
maximize a robust measure of spread to obtain 
consecutive directions on which the data points are 
projected,
see~\cite{Hubert:RAPCA,Croux:ProjRPCA}.

The ROBPCA~\cite{Hubert:ROBPCA} approach is a hybrid, 
which combines ideas of projection pursuit and robust 
covariance estimation. 
The projection pursuit part is used for the initial 
dimension reduction. Some ideas based on the MCD 
estimator are then applied to this lower-dimensional 
data space.

\begin{figure}[H]
\begin{center}
\includegraphics[width=1.0\textwidth]
                {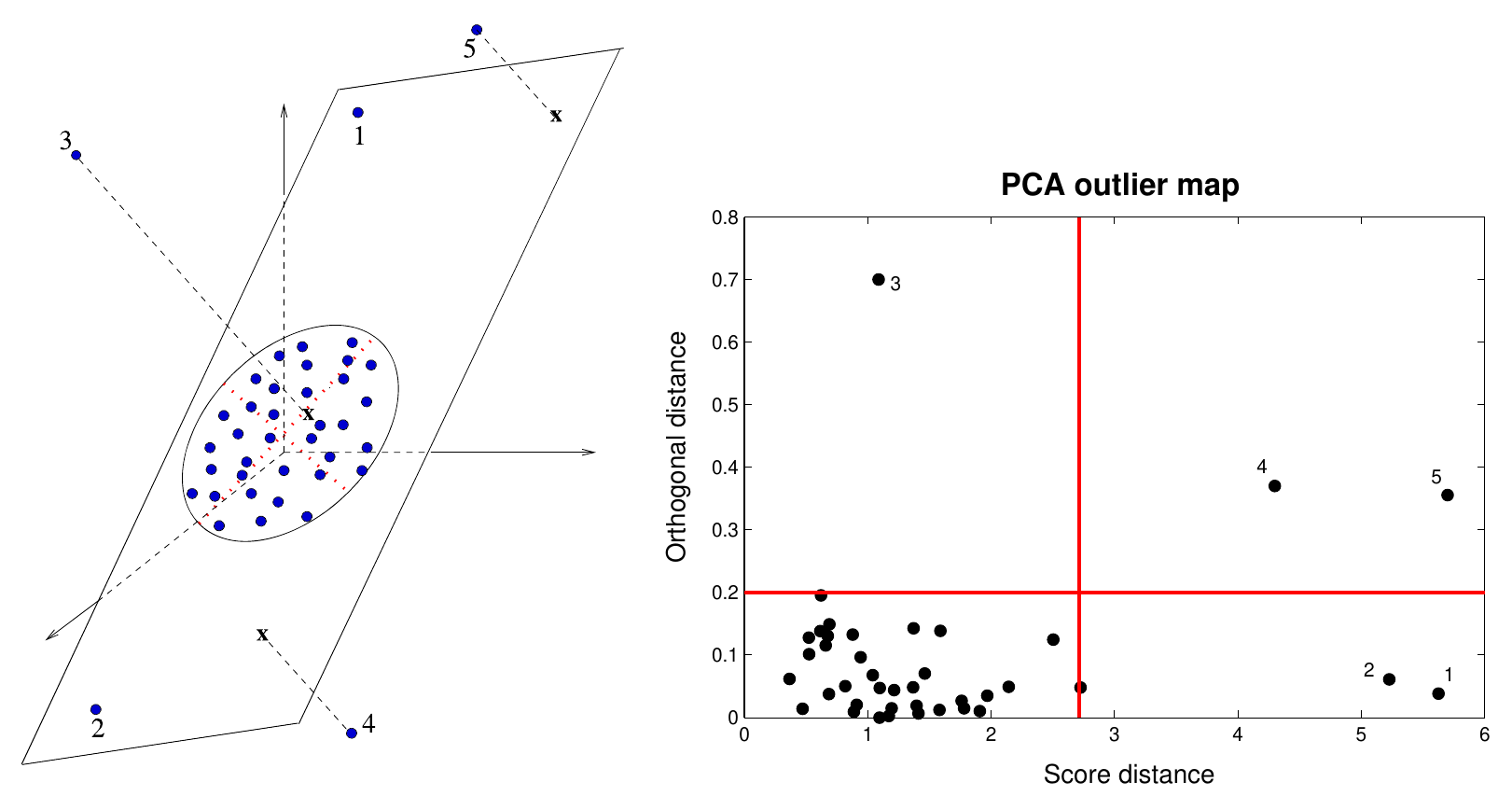}
\caption{Illustration of PCA: (left) types of outliers;
				 (right) outlier map: plot of orthogonal 
				 distances versus score distances.}
\label{pca_stylized_OM} 
\end{center}
\end{figure}

In order to diagnose outliers we can draw an outlier 
map for PCA~\cite{Hubert:ROBPCA}, similar to the
outlier map for regression in the previous section.
A stylized example of such a PCA outlier map is shown in
the right panel of Figure \ref{pca_stylized_OM}, which
corresponds to the three-dimensional data in the left 
panel which is fitted by two principal components.
On the vertical axis of the PCA outlier map we find the 
{\it orthogonal distance} of each data point to the PCA 
subspace. This is just the Euclidean distance of the
data point to its projection. The orthogonal distance
is highest for the points 3, 4, and 5 in the example.
On the horizontal axis we see the {\it score distance} 
of each data point, which is just the robust distance 
\eqref{eq:RD} of its projection relative to all the
projected data points. The score distance is rather high
for the points 1, 2, 4, and 5 in the figure.

By combining both distance measures the outlier map 
allows to distinguish between four types of data points.
{\it Regular observations} have both a small orthogonal 
distance and a small score distance. 
Points with a high score distance but a small orthogonal 
distance, such as points 1 and 2 in 
Figure \ref{pca_stylized_OM}, are called {\it good 
leverage points} as they can improve the accuracy of the 
fitted PCA subspace. 
{\it Orthogonal outliers} have a large orthogonal 
distance but a small score distance, like point 3.
{\it Bad leverage points} have both a large orthogonal 
distance and a large score distance, like points 4 and 5.
They lie far from the space spanned by the robust 
principal components, and after projection on that space
they lie far from most of the other projected data. 
They are called `bad' because they typically they have a 
large influence on classical PCA, as the eigenvectors 
will be tilted towards them.

\begin{figure}[H]
\begin{center}
%\vskip0.2cm
\includegraphics[width=1.0\textwidth]
                {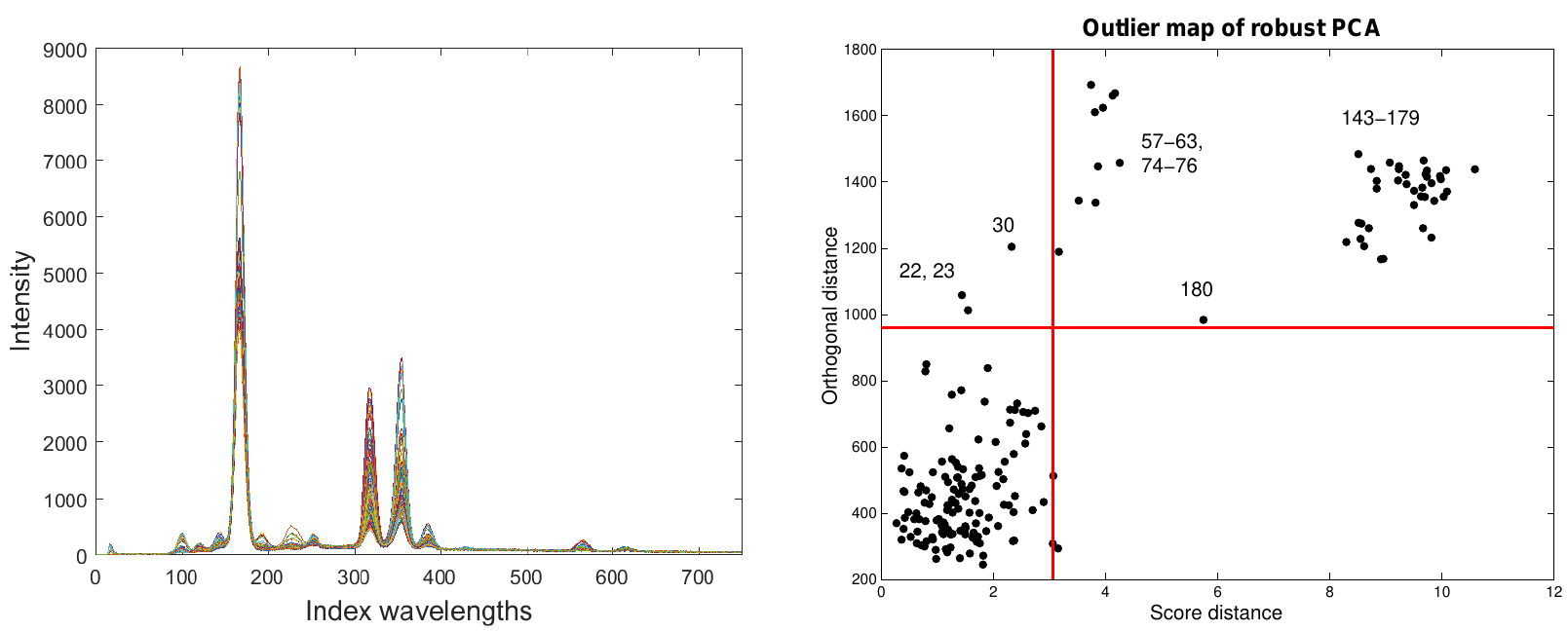}
\caption{Glass data: (left) spectra;
				 (right) outlier map.}
\label{pca_glass} 
\end{center}
\vskip-0.5cm
\end{figure}

As a real example we take the glass 
data~\cite{Lemberge:glass} consisting of spectra of
180 archaeological glass vessels from the 16th--17th 
centuries. They have 750 wavelengths so $d > n$.
The spectra are shown in Figure~\ref{pca_glass} with
their outlier map based on ROBPCA, which clearly indicates
a substantial number of bad leverage points and several
orthogonal outliers. An analogous
plot based on classical PCA (not shown) did not reveal 
the outliers, because they tilted the PCA subspace toward
them. Also the plots of the first few principal 
components looked quite different.

Other proposals for robust PCA include
spherical PCA~\cite{Locantore:Funcdata} which 
first projects the data onto a sphere with a 
robust center, and then applies PCA to 
these projected data. To obtain sparse loadings, a 
robust sparse PCA method is proposed 
in~\cite{Hubert:ROSPCA}.
When linear models are not appropriate, one may use
support vector machines (SVM) which are powerful tools 
for handling nonlinear structures~\cite{Steinwart:Book}.
A kernelized version of ROBPCA (KROBPCA) 
is introduced in~\cite{Debruyne:RobustKernelPCA}.
For a review of robust versions of principal 
component regression and partial least squares
see~\cite{Hubert:ReviewHighBreakdown}.

\section*{\sffamily \Large OTHER MODELS} 

\subsection*{\sffamily \large Classification}
The goal of classification, also known as discriminant 
analysis or supervised learning, is to obtain rules 
that describe the separation between known groups 
$G_j$ of $d$-dimensional data points, with an eye 
toward assigning new data points to one of the groups. 
We write $p_j$ for the membership probability, i.e.\ the 
probability for any observation to come from $G_j$.

For low-dimensional data, a popular classification 
rule results from maximizing the Bayes posterior 
probability. At gaussian distributions this yields 
quadratic discriminant analysis (QDA), i.e. choosing 
the $j$ for which $\bx$ has the highest 
quadratic score $d_j^Q(\bx)$ given by
\begin{equation}
  d_j^Q(\boldsymbol{x}) =
  - \frac{1}{2}\mbox{ln}|\boldsymbol{\Sigma}_j| -
  \frac{1}{2}(\boldsymbol{x} - \boldsymbol{\mu}_j)^T
	\bSigma_j^{-1}(\boldsymbol{x} - \boldsymbol{\mu}_j)+
	\mbox{ln}(p_j)\;.
  \label{eq:quadrdiscrscores}
\end{equation}
When all the covariance matrices are assumed to be equal, 
these scores can be simplified to
\begin{equation}\label{eq:lineardiscrscores}
  d_j^\mathit{L}(\bx)= \bmu^T_j \bSigma^{-1}\bx
  - \frac{1}{2}\bmu^T_j\bSigma^{-1}\bmu_j + \ln(p_j)
\end{equation}
where $\bSigma$ is the common covariance matrix, yielding
linear discriminant analysis (LDA). Robust classification 
rules can be obtained by replacing the classical covariance 
matrices by robust alternatives such as the MCD estimator 
or S-estimators, as
in~\cite{Hawkins:Discrim,He:Discrim,Croux:Discrim,
Hubert:Discrim}.

When the data are high-dimensional, this approach cannot 
be applied because the robust covariance estimators are 
no longer computable. 
One approach is to first apply robust PCA to the entire 
data set. Alternatively, one can also apply a PCA method 
to each group separately. This is the idea behind the 
SIMCA (Soft Independent Modeling of Class Analogy) 
method~\cite{Wold:PatRecogn}. 
A robustification of SIMCA is obtained by first applying 
robust PCA to each group, and then constructing a 
classification rule for new observations based on their 
orthogonal distance to each subspace and their score 
distance within each subspace~\cite{VandenBranden:RSIMCA}.

An SVM classifier with an unbounded kernel, e.g.\ 
a linear kernel, 
is not robust and suffers the same problems as 
traditional linear classifiers. But when a bounded kernel 
is used, the resulting non-linear SVM classification 
handles outliers quite well~\cite{Steinwart:Book}. 
As an alternative, one can apply KROBPCA combined with 
LDA to the scores~\cite{Debruyne:RobustKernelPCA}.

\subsection*{\sffamily \large Clustering}
Cluster analysis (also known as unsupervised learning) 
is an important methodology when handling large 
data sets. It searches for homogeneous groups in the
 data, which afterwards may be analyzed separately. 
Partitioning (non-hierarchical) clustering methods 
search for the best clustering in $k$ groups.

For spherical clusters, the most popular method is 
$k$-means which minimizes the sum of the squared
 Euclidean distances of the observations to the mean 
of their group~\cite{McQueen:kmeans}. 
This method is not robust as it uses averages. 
To overcome this problem, one of the first robust
proposals was the Partitioning Around Medoids 
method~\cite{Kaufman:cluster}. 
It searches for $k$ observations (called medoids) such 
that the sum of the unsquared distances of the 
observations to the medoid of their group is minimized. 
The CLARA algorithm~\cite{Kaufman:cluster} implemented 
this method for large datasets, and was extended to 
CLARANS~\cite{Ng:CLARANS} for spatial data mining.

Later on the more robust trimmed $k$-means method has 
been proposed~\cite{Cuesta:kmeans}, inspired by the 
trimming ideas in the MCD and the LTS. It searches for 
the $h$-subset (with $h$ as in the definition of MCD) 
such that the sum of the squared distances of 
the observations to the mean of their group is 
minimized. Consequently, not all observations need to 
be classified, as $n-h$ cases can be left unassigned. 
To perform the trimmed $k$-means clustering an iterative 
algorithm~\cite{Garcia:kmeans} has been developed, using 
C-steps like those in the FAST-MCD algorithm. 
For non-spherical clusters, constrained maximum 
likelihood 
approaches~\cite{Gallegos:cluster,Garcia:tclust} were 
developed.

\subsection*{\sffamily \large Functional data}
In functional data analysis, the cases are not data 
points but functions. 
A functional data set typically consists of $n$ curves 
observed on a set of gridpoints $t_1,\ldots,t_T$\;.
These curves can have smoothness properties, numerical 
derivatives and so on.
Standard references on functional data are the
books~\cite{Ramsay:BookFDA,Ferraty:BookFDA}.
A functional data set can be analyzed by principal 
components, for which robust methods are 
available~\cite{Boente:FPCA}. To classify functional 
data, a recent approach is presented 
in~\cite{Hubert:MFC}.

The literature on outlier detection in functional data 
is rather young, and several graphical tools have been
developed~\cite{Hyndman:Bagplot,Sun:FuncBoxplots,
Arribas:Outliergram}, mainly for univariate functions.
In~\cite{Hubert:MFOD} also multivariate functions are
discussed and a taxonomy of functional outliers is
set up, with on the one hand functions that are outlying 
on most of their domain, such as shift and magnitude 
outliers as well as shape outliers, and on 
the other hand isolated outliers which are only outlying
on a small part of their domain. 
The proposed heatmap and functional outlier map
are tools to flag outliers and detect their type.
This work is expanded in~\cite{DOIV} to functional data 
with bivariate domains, such as images and video.

\subsection*{\sffamily \large Other applications}
Robust statistics has many other uses apart from 
outlier detection. 
For instance, robust estimation can be used in automated
settings such as computer 
vision~\cite{Meer:RobCV,Stewart:CV}.
Another aspect is statistical inference, such as the 
construction of robust hypothesis tests, $p$-values, 
confidence intervals, and model selection
(e.g.\ variable selection in regression). This aspect
is studied in~\cite{Hampel:IFapproach,Maronna:RobStat}
and in other works they reference.

\section*{\sffamily \Large DETECTING OUTLYING CELLS} 
Until recently people have always considered outliers 
to be cases (data points), i.e.\ rows of 
the $n \times d$ data matrix $\bX$.
But recently the realization has grown that this
paradigm is no longer sufficient for the
high-dimensional data sets we are often faced with
nowadays. 
Typically most data cells (entries) in a row are 
regular and a few cells are anomalous.
The first paper to formulate the cellwise paradigm 
was~\cite{Alqallaf}, which showed how such outlying
cells propagate in computations. In more than a few
dimensions, even a small percentage of outlying cells
can spoil a large percentage of rows. This is fatal
for rowwise robust methods, which require at least
50\% of the rows to be clean.

Detecting cellwise outliers is a hard problem, since
the outlyingness of a cell depends on the relation of
its column to the other columns of the data, and on
the values of the other cells in its row (some of 
which may be outlying themselves).
The {\it DetectDeviatingCells}~\cite{DDC} algorithm
addresses these issues, and apart from flagging cells 
it also provides a graphical output called a cellmap.

\newpage
\begin{figure}[ht]
\begin{center}
\includegraphics[keepaspectratio=true,scale=0.48]
                {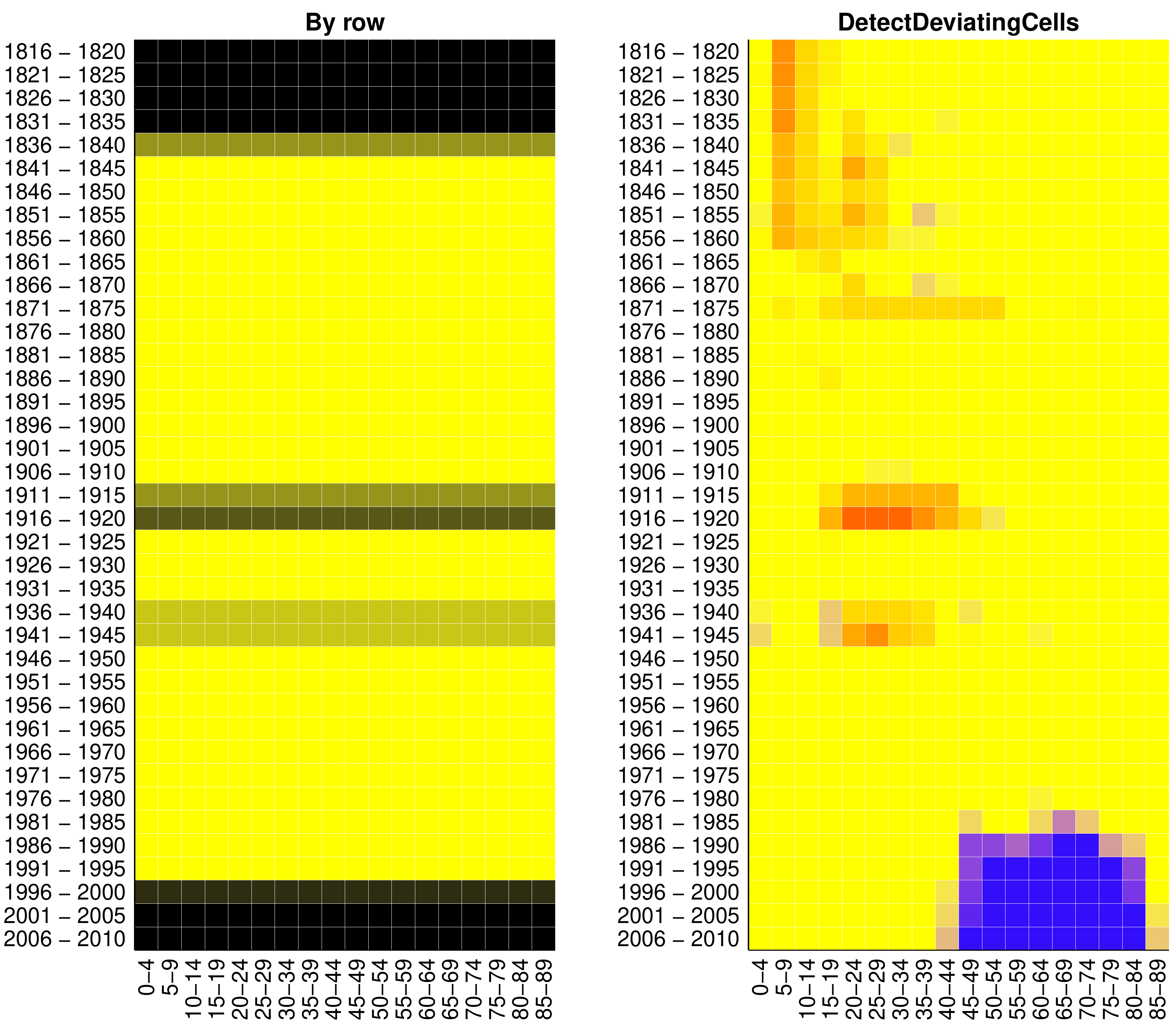}				
\caption{Male mortality in France in 1816--2010: (left)
         detecting outlying rows by a robust PCA method; 
				 (right) detecting outlying cells by
				 {\it DetectDeviatingCells}. After the analysis, the
				 cells were grouped in blocks of $5 \times 5$
				 for visibility.}
\label{DDC_mortality} 
\end{center}
\end{figure}

As an example we consider the mortality by age for 
males in France from 1816 to 2010, obtained from
\url{http://www.mortality.org}\;.
Each row corresponds to the mortalities in a given 
calendar year.
The left panel in Figure \ref{DDC_mortality} shows
the result of the ROBPCA method described in the 
section on principal components. 
Outlying rows are shown in black and regular rows 
in yellow.
The analysis was carried out on the data set with the
individual years and the individual ages, but as this
resolution would be too high to fit on the page we have
combined the cells into $5 \times 5$ blocks afterward.
The combination of some black rows with some yellow 
ones has led to gray blocks.
We can see that there were outlying rows in the early
years, the most recent years, and during two periods
in between.
Note that a black row doesn't provide information
about its cells.

By contrast, the result of {\it DetectDeviatingCells}
in the right panel in Figure~\ref{DDC_mortality}
identifies a lot more information.
Cells with higher values than predicted are shown in
red, and those with lower values in blue, after which
the colors were averaged in the $5 \times 5$ blocks.
The outlying early years saw a high infant mortality.
During the Prussian war and both world wars there was 
a higher mortality among young adult men.
And in recent years mortality among middle-aged and 
older men has decreased substantially, perhaps due to
medical advances.

\begin{figure}[H]
\begin{center}
\includegraphics[keepaspectratio=true,scale=0.41]
                {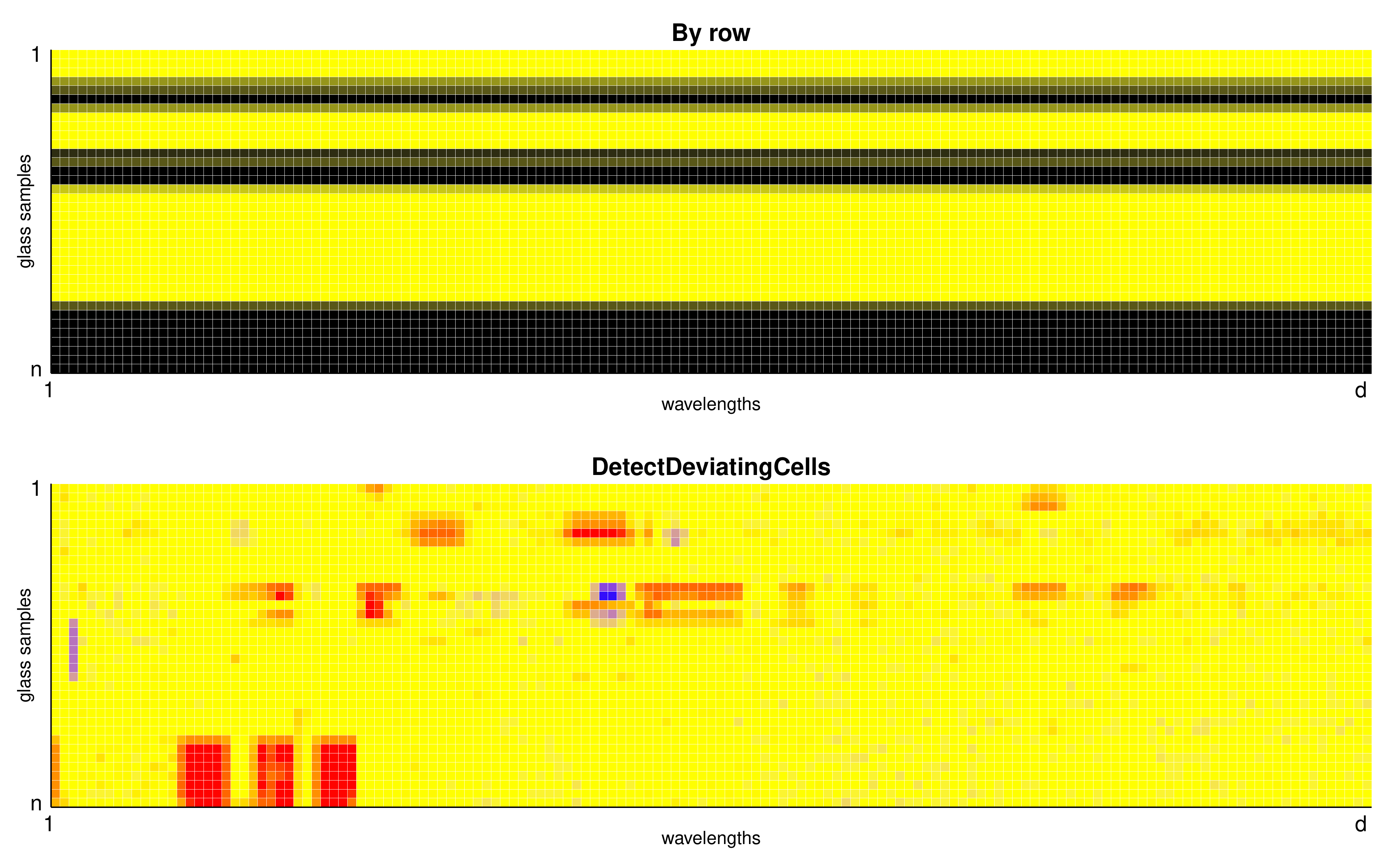}	
\caption{Cell map of the glass data. The positions of
         the deviating cells reveal the chemical
				 contaminants.}
\label{DDC_glass} 
\end{center}
\end{figure}

We also return to the glass data from the section on 
PCA.
The top panel in Figure~\ref{DDC_glass} shows the 
rows detected by the ROBPCA method.
The lower panel is the cell map obtained by 
{\it DetectDeviatingCells} on this $180 \times 750$ 
dataset.
After the analysis, the cells were again grouped in 
$5 \times 5$ blocks.
We now see clearly which parts of each spectrum are
higher/lower than predicted.
The wavelengths of these deviating cells reveal the
chemical elements responsible.

Ideally, after running{\it DetectDeviatingCells} the
user can look at the deviating cells and whether their
values are higher or lower than predicted, and make 
sense of what is going on. 
This may lead to a better understanding of the
data pattern, to changes in the way the data
are collected/measured, to dropping certain
rows or columns, to transforming variables,
to changing the model, and so on.
If the data set is too large for visual inspection
of the results, or the analysis is
automated, the deviating cells can be set to missing
after which the data set is treated by a method
appropriate for incomplete data.
A good rowwise robust method of this type 
is~\cite{Agostinelli}.

\section*{\sffamily \Large SOFTWARE AVAILABILITY} 
All the examples in this paper were produced with the
free software R~\cite{R:core}. 
The publicly available CRAN package 
\texttt{robustbase} contains Qn, covMcd, ltsReg, 
and lmrob, whereas \texttt{rrcov} has many robust 
covariance estimators, robust principal components, 
and robust LDA and QDA classification.
ROBPCA and its extensions are available in
\texttt{rospca} and robust SIMCA in \texttt{rrcovHD}.
Robust clustering can be performed with the 
\texttt{cluster} and \texttt{tclust} packages.
The package \texttt{mrfDepth} contains tools for 
functional data and \texttt{cellWise} provides 
cellwise outlier detection and cellmaps.

Matlab functions for many of these methods are 
available in the LIBRA 
toolbox~\cite{Verboven:Toolbox,Verboven:WIRE-LIBRA},
which can be downloaded from 
\url{http://wis.kuleuven.be/stat/robust}\;.

The MCD and LTS methods are also built into S-PLUS 
as well as SAS (version 11 or higher) and SAS/IML 
(version 7 or higher).

\section*{\sffamily \Large CONCLUSIONS}
We have surveyed the utility of robust statistical methods
and their algorithms for detecting anomalous data.
These methods were illustrated on real data, in frameworks
ranging from covariance matrices,
the linear regression model and principal component analysis,
with references to methods for many other tasks such as 
supervised and unsupervised classification as well as the
analysis of functional data. For high-dimensional data,
sparse and regularized robust methods were developed
recently.

We have described methods to detect anomalous cases 
(rowwise outliers) but also newer work on the detection 
of anomalous data cells (cellwise outliers). 
An important topic for future research is to further improve 
the efficiency of the robust methodologies, in terms of 
both predictive accuracy and computational cost.

\end{document}